\documentclass{article}
\usepackage{spconf,amsmath,graphicx}
\usepackage{booktabs}
\usepackage{multirow}
\usepackage{xcolor}
\usepackage{float}
\usepackage{url}
\usepackage{hyperref}
\usepackage{graphicx}

\usepackage{tikz}
\usepackage{url}
\usepackage{hyperref}
\usepackage{eso-pic}

\newcommand{\addpublisheddetails}[3]{%
    \AddToShipoutPictureBG*{%
        \AtPageUpperLeft{%
            \begin{tikzpicture}[remember picture,overlay]
                \node[anchor=north west, fill=yellow!20, text width=\dimexpr\paperwidth-1cm, align=left, font=\sffamily\small, inner sep=10pt, minimum height=3cm] 
                at ([xshift=0cm, yshift=0cm]current page.north west) {%
                    \begin{center}
                        © 2024 IEEE. Personal use of this material is permitted. Permission from IEEE must be obtained for all other uses.
                    \end{center}
                    \textbf{Published Version:} {\footnotesize #1} \\[0.3em]
                    \textbf{Citation:} {\footnotesize #2} \\[0.3em]
                    \textbf{Access:} {\footnotesize \href{#3}{\texttt{#3}}}
                };
            \end{tikzpicture}
        }
    }
}
\addpublisheddetails{2024 IEEE Spoken Language Technology Workshop (SLT)}
{La Quatra, M., Salerno, V. M., Tsao, Y., Siniscalchi, S. M., ``FlanEC: Exploring Flan-T5 for Post-ASR Error Correction,'' SLT, 2024, pp. 608-615.}
{https://doi.org/10.1109/SLT61566.2024.10832257}

\newcommand{\enna}{$^{\star}$}
\newcommand{\acsinica}{$^{\dagger}$}
\newcommand{\unipa}{$^{\ddagger}$}

\newcommand{\modelname}{FlanEC}
\newcommand{\modelnamecdf}{FlanEC-CD}
\newcommand{\modelnamesdf}{FlanEC-SD}

\definecolor{lightblue}{rgb}{0.8, 0.9, 1}
\definecolor{lightred}{rgb}{1, 0.8, 0.8}

\title{FLANEC: Exploring Flan-T5 for Post-ASR Error Correction}
\name{
    Moreno La Quatra\enna, 
    Valerio Mario Salerno\enna, 
    Yu Tsao\acsinica,
    \textit{Sabato Marco Siniscalchi}\unipa
}
\vspace{5mm}
\address{
    {\enna Kore University of Enna}, 
    {\acsinica Academia Sinica},
    {\unipa Università degli Studi di Palermo}
}

\begin{document}
%
\maketitle
\begin{abstract}
In this paper, we present an encoder-decoder model leveraging Flan-T5 for post-Automatic Speech Recognition (ASR) Generative Speech Error Correction (GenSEC), and we refer to it as FlanEC. We explore its application within the GenSEC framework to enhance ASR outputs by mapping n-best hypotheses into a single output sentence. By utilizing n-best lists from ASR models, we aim to improve the linguistic correctness, accuracy, and grammaticality of final ASR transcriptions. Specifically, we investigate whether scaling the training data and incorporating diverse datasets can lead to significant improvements in post-ASR error correction. We evaluate FlanEC using the HyPoradise dataset, providing a comprehensive analysis of the model's effectiveness in this domain. Furthermore, we assess the proposed approach under different settings to evaluate model scalability and efficiency, offering valuable insights into the potential of instruction-tuned encoder-decoder models for this task.
\end{abstract}
\section{Introduction}
\label{sec:intro}

Generative Speech Error Correction (GenSEC) focuses on correcting linguistic errors in transcriptions of spoken language. 
In Automatic Speech Recognition (ASR) systems, GenSEC acts as an effective post-processing step to enhance the accuracy \cite{toshniwal2018comparison, kannan2018analysis}. 
This step is crucial when ASR systems are used as inputs to downstream applications such as machine translation or information retrieval, where linguistic correctness is essential for maintaining the quality and reliability of the final output \cite{sodhi2021mondegreen}.
While modern end-to-end ASR systems have significantly improved transcription accuracy, their primary focus on acoustic modeling often results in neglecting grammatical error correction \cite{radford2023robust}.
These systems excel at capturing acoustic cues and accurately transcribing spoken words.
However, the inherent challenges of speech-to-text conversion, including variability in speech patterns, accents, and background noise, can lead to disfluencies, repetitions, and other speech irregularities that compromise the correctness of the final text.
Therefore, ASR outputs may require additional post-processing to address linguistic errors and improve the overall quality of the transcription.
The primary objective of GenSEC is to correct these errors by exploiting the context provided by $n$-best hypotheses generated by ASR models and leveraging external linguistic knowledge to produce a more accurate transcription.
An $n$-best list, in this context, is an ordered list of the top $n$ possible transcriptions generated by an ASR model, each with its associated confidence score.
Recent advancements in natural language processing (NLP) have shown that large language models (LLMs) can be effectively adapted for various tasks.
In this context, the Flan-T5 model \cite{chung2024scaling} has demonstrated strong performance through instruction tuning, a process that specifically trains a single model to perform a diverse set of tasks by following specific instructions.
This method significantly enhances the model's generalization capabilities, enabling it to adapt to specific tasks with minimal adjustments.
By leveraging Flan-T5's performance in understanding and generating natural language \cite{raheja-etal-2023-coedit}, we aim to address the inherent challenges of post-ASR correction.  

\noindent
This paper presents our system for Task 1 of the GenSEC Grand Challenge, which focuses on post-ASR correction using large language models (LLMs). We propose \modelname{} (\textit{Flan}-T5 for \textit{E}rror \textit{C}orrection), an encoder-decoder model built on Flan-T5 \cite{chung2024scaling}. 
Our contributions are as follows:
\begin{itemize}
    \item We introduce \modelname{}, an encoder-decoder model specifically instructed for GenSEC, trained at various scales from 250 million to 3 billion parameters.
    \item We evaluate \modelname{} using the HyPoradise dataset \cite{chen2024hyporadise}, providing a comprehensive overview of its capabilities across different ASR domains.
    \item We explore the impact of scaling training data across all subsets of the HyPoradise dataset to assess whether training a single model on diverse domains can enhance performance across different subsets.
    \item We investigate the effectiveness of efficient model adaptation (i.e., LoRA \cite{hu2021lora}) versus full fine-tuning (FT) to assess to which extent it can provide comparable performance to full fine-tuning in our task.
\end{itemize}

\section{Post-ASR Error Correction}
\label{sec:GEC}

Correcting linguistic errors is a fundamental challenge in NLP and has traditionally focused on written text, with significant advancements documented over the years \cite{ng-etal-2014-conll}.
LLMs have shown remarkable capabilities in understanding and generating natural language, making them well-suited for GenSEC tasks \cite{touvron2023llama, brown2020language, chung2024scaling}. 
By integrating general knowledge and linguistic context, LLMs as a post-processing step can leverage the strengths of both ASR and NLP systems to enhance transcription accuracy.

\subsection{Generative Speech Error Correction}
\label{subsec:gensec}
GenSEC is an emerging field dedicated to enhancing the accuracy of ASR outputs by leveraging text-only information. 
It aims to generate a single correct transcription given one or more ASR hypotheses, each potentially accompanied by a confidence score.
Previous works have demonstrated that integrating language models as an additional step on top of ASR models can significantly improve the quality of ASR transcriptions \cite{toshniwal2018comparison} with RNN-based LMs showing better performance than n-gram LMs \cite{kannan2018analysis}.
Recent advancements in LLMs have further corroborated this hypothesis, showing that specific prompts, together with fine-tuning, can enhance the quality of the final transcriptions \cite{yang2023generative}.
By combining the technical strengths of ASR systems with the advanced linguistic processing power of LLMs, GenSEC can lead to transcriptions that are more accurate and linguistically coherent.

In this context, the HyPoradise dataset \cite{chen2024hyporadise} serves as a comprehensive benchmark for post-ASR error correction tasks, spanning several ASR domains to assess GenSEC models under various conditions.
The dataset provides a standardized framework for assessing the effectiveness of different approaches to post-ASR error correction, emphasizing the linguistic aspects of the task without requiring additional acoustic modeling or alignment. 
Each spoken utterance within the dataset is accompanied by an $n$-best list of hypotheses generated by an ASR model (e.g., \cite{radford2023robust}), along with the corresponding reference transcription. 
Our work aims at exploring the potential of encoder-decoder language models to enhance GenSEC tasks using the HyPoradise dataset.

\subsection{LLMs in GenSEC}

Instruction tuning has highlighted the adaptability of LLMs, enabling them to effectively shift between different tasks by leveraging explicit instructions \cite{mishra-etal-2022-cross}. 
This approach enhances the model's ability to handle diverse NLP challenges efficiently.
In-context learning \cite{dong2022survey, wei2022chain, wang2024can} is a notable capability of LLMs, it allows models to learn directly from examples provided within a prompt, adapting to perform specific tasks using only the information given in the input. 
The emerging abilities of LLMs in this domain demonstrate their potential for handling a wide range of linguistic tasks \cite{wei2022emergent}.
In the context of GenSEC, LLMs are used to generate corrections for ASR hypotheses, simultaneously leveraging the context provided by the ASR outputs and the linguistic knowledge embedded in the model \cite{hu2024large,10447938}.
Recent studies \cite{mosbach-etal-2023-shot}, however, have demonstrated that adapting model parameters outperforms in-context learning across various model scales. 
Thus, we focus on using an open-source model to tune parameters for post-ASR error correction.

While most LLMs are decoder-only, we focus on an encoder-decoder architecture that is open-source, available in various sizes, and capable of effectively following instructions. 
Flan-T5 \cite{chung2024scaling} meets these criteria and has been pre-trained to solve 1,836 diverse tasks, significantly improving T5 \cite{roberts2019exploring} ability to generalize across different, even unseen, instructions.
These characteristics make Flan-T5 particularly well-suited for GenSEC tasks, where the model must adapt to a range of linguistic contexts to generate accurate transcriptions by exploiting the context provided through ASR hypotheses.
Flan-T5 has demonstrated superior performance in handling instruction-based tasks, even when compared to decoder-only LLMs of similar parameter scale, as evidenced by its results in the InstructEval benchmark \cite{chia2023instructeval}.

\section{Flanec}
\label{sec:model}

The proposed \modelname{} is an encoder-decoder model built on Flan-T5, specifically instructed for GenSEC tasks.
We define a specific instruction template 
and use it to create \modelname{} using HyPoradise dataset and leveraging the $n$-best lists generated by an ASR model. 

\subsection{Task prompt}

The prompt for \modelname{} is designed to instruct the model to generate a single correct transcription from a set of ASR hypotheses. To this end, the HyPoradise dataset, denoted as $D$, is used. 
It is composed of different subsets $D_1, D_2, \ldots, D_m$. A sample $s_{ij} \in D_i$, namely  the  $j$-th sample in the subset $D_i$, consists of (i) the $n$-best list of ASR hypotheses $H_{ij} = \{h_{ij1}, h_{ij2}, \ldots, h_{ijn}\}$, and (ii) the corresponding reference transcription $r_{ij}$.
\modelname{} is trained to map $H_{ij} \rightarrow r_{ij}$ by generating a single output written sentence that aligns with the reference transcription.

The task template is designed to be consistent with the original Flan-T5 approach, using natural language instructions without specific special tokens or tags \cite{chung2024scaling}. The following prompt is intended to guide the model in generating the correct transcription from the $n$-best list:

\begin{tabbing}
    \hspace{1.3cm} \= \hspace{8cm} \= \kill
    \textbf{Input}: \> Generate the correct transcription for the following \+ \\ 
     n-best list of ASR hypotheses: \- \\
    \> - $h_{ij1}$ \\
    \> - $h_{ij2}$ \\
    \> - \ldots \\
    \> - $h_{ijn}$ \\
    \textbf{Output}: \> $r_{ij}$
\end{tabbing}

\noindent
Where $i$ denotes the subset index and $j$ denotes the sample index within the subset. 
In our experiments, we use $n=5$ for the list of hypotheses, as \cite{ma23e_interspeech} demonstrates this setting yields the best results.
The model is trained to generate the correct transcription $r_{ij}$ by exploiting the context provided in the list of hypotheses. 
To fully leverage the capabilities of Flan-T5, we did not apply any additional constraints on the decoding process, allowing the model to generate the output freely based on the input context. 
We speculate that a sufficient amount of training data, combined with the instruction-following capabilities of Flan-T5 and the context provided by the ASR hypotheses, should be enough to prevent the model from deviating from the desired output without limiting its ability to generate accurate transcriptions. 
Supporting this approach, our analysis presented in Table \ref{tab:stats_tokens} shows that, on average across datasets, more than 4.5\% of tokens in the reference transcriptions are not present in any of the $n$-best hypotheses.
Notably, in the CV-Accent dataset, more than 8\% of tokens in the reference transcriptions are new, underscoring the importance of allowing the model to generate beyond the provided hypotheses.
Furthermore, the analysis reveals that a significant percentage of reference sentences do not match any of the sentences in the $n$-best lists. 
This indicates that limiting the decoding process to only the tokens or sentences that appear at least once in the original prompt, as suggested in \cite{ma23e_interspeech}, may restrict the model's correction capabilities and its use of internal linguistic knowledge.

\begin{table}[]
\centering
\resizebox{\columnwidth}{!}{%
\begin{tabular}{@{}c|ccc|ccl@{}}
\toprule
\multirow{2}{*}{Dataset} & \multicolumn{3}{c|}{Training} & \multicolumn{3}{c}{Test}                      \\
                         & Avg. NT  & \% NT    & \% NS   & Avg. NT & \% NT   & \% NS \\ \midrule
WSJ                      & 0.83     & 3.19   & 40.74 & 0.46    & 1.86  & 25.00                  \\
ATIS                     & 0.33     & 1.60   & 25.81 & 0.32    & 1.75  & 26.08                   \\
CHiME-4                  & 2.28     & 9.83   & 61.06 & 0.93    & 4.21  & 47.65                   \\
Tedlium-3                & 0.49     & 2.26   & 42.80 & 0.30    & 1.07  & 34.11                   \\
CV-accent                & 1.40     & 8.47   & 51.36 & 1.46    & 8.74  & 51.90                   \\
SwitchBoard              & 1.32     & 6.04   & 71.08 & 1.24    & 5.80  & 70.95                   \\
LRS2                     & 0.27     & 2.69   & 29.86 & 0.29    & 2.92  & 29.97                   \\
CORAAL                   & 6.83     & 11.54  & 91.62 & 7.31    & 12.47 & 94.12                   \\ \midrule
Overall                  & 0.98     & 4.79   & 47.44 & 0.90    & 4.51  & 44.90                   \\ \bottomrule
\end{tabular}%
}
\caption{Average and percentage of New Tokens (NT) present in reference transcriptions but absent in any of the $n$-best hypotheses, and the percentage of New Sentences (NS) where the reference does not match any $n$-best proposal, for training and test sets of each HyPoradise subset, and overall.}
\label{tab:stats_tokens}
\end{table}

\subsection{Fine-tuning strategy}

In line with Flan-T5, we evaluate two training settings:
\begin{itemize}
    \item \textit{SD}: Single Dataset, where we train a separate model on each subset of the HyPoradise dataset (e.g., ATIS, Tedlium3, etc.).
    \item \textit{CD}: Cumulative Dataset, where we train a single model on the combined subsets of the HyPoradise dataset.
\end{itemize}

\noindent
In \textit{SD}, we train a separate model on each $D_i \in D$, whereas in \textit{CD}, we train a single model on the union of all subsets, $\bigcup_{i=1}^{m} D_i$.
While \textit{SD} has been the standard approach for training GenSEC models \cite{chen2024hyporadise}, in \textit{CD} settings, we aim to create an \textit{omnibus} model capable of generalizing across different ASR domains. 
Our goal is to determine if \modelname{} can generalize by leveraging multiple domains, similar to how Flan-T5 generalizes across different tasks. 
By combining diverse datasets such as ATIS (Airline Travel Information Systems), Tedlium3 (transcripts from TED talks), Switchboard (telephone conversation transcripts), and other training subsets in the HyPoradise dataset, we create a more varied training set. 
This approach utilizes the diversity of the training data to enhance the model's ability to correct errors across various ASR subsets. 
We hypothesize that training a single model on these diverse datasets will lead to better generalization and improved performance across multiple ASR domains. 
By learning from the combined data, the model can benefit from shared knowledge, potentially enhancing its overall performance.

To explore efficient fine-tuning methods, we also employ LoRA \cite{hu2021lora}, which has shown promising results in adapting LLMs to specific tasks. 
LoRA involves training adapters on top of a pre-trained model, allowing for cost-effective fine-tuning without altering the original model's parameters. These adapters consist of additional, task-specific parameters trained to adapt the model to the target task. 
While LoRA is computationally efficient, it may not always match the performance of full fine-tuning (FT). 
We investigate this trade-off in the experimental section by training \modelname{} using both LoRA and FT and comparatively evaluating their performance.

\section{Experiments}
\label{sec:experiments}

We evaluate \modelname{} on the HyPoradise dataset, focusing on post-ASR error correction across different ASR domains. 
We adapt the model at various scales, ranging from 250 million to 3 billion parameters, to assess the impact of scaling model parameters on GenSEC performance. 
We compare the model's performance under different training settings, including \textit{SD} and \textit{CD}, to evaluate its generalization capabilities across diverse ASR domains. 
Additionally, we explore the effectiveness of LoRA versus full fine-tuning (FT) to assess the model's scalability and efficiency in the GenSEC task\footnote{Code to reproduce the experiments is available at: \url{https://github.com/MorenoLaQuatra/FlanEC}}.
We refer to \textit{adaptation} as the process of updating model parameters, whether through the LoRA method or full fine-tuning.

\subsection{HyPoradise Dataset}
\label{subsec:dataset}

Task 1 of the GenSEC Grand Challenge focuses on post-ASR error correction using the HyPoradise dataset \cite{chen2024hyporadise}. 
This  benchmark is designed to explore LLMs' capabilities in correcting ASR errors across various domains. 
It includes eight different subsets, each representing a distinct domain:

\begin{enumerate}
    \item \textit{WSJ} \cite{wsj_dataset}: Business news and financial data readings.
    \item \textit{ATIS} \cite{atis_dataset}: Airline Travel Information Systems queries.
    \item \textit{CHiME-4} \cite{chime4_dataset}: Noisy speech data from different environments, part of the CHiME-4 challenge.
    \item \textit{Tedlium-3} \cite{tedlium3_dataset}: Transcripts from TED talks.
    \item \textit{CV-accent} \cite{cv_dataset}: Accented speech data from the English portion of the Common Voice dataset.
    \item \textit{SwitchBoard} \cite{switchboard_dataset}: Telephone conversation transcripts.
    \item \textit{LRS2} \cite{lrs2_dataset}: Audio clips from BBC programs.
    \item \textit{CORAAL} \cite{coraal_dataset}: Interviews containing accented speech from regional varieties of African American English.
\end{enumerate}

\noindent
The diversity of the data ensures that models are evaluated across a wide range of ASR domains, providing a comprehensive assessment of their capabilities. 
Each subset contains both training and test data, comprising spoken utterances, $n$-best lists generated by an ASR model, and reference transcriptions. 
In our experiments, we train \modelnamecdf{} on $\bigcup_{i=1}^{m} D_i$ and evaluate its performance on the test subsets of the HyPoradise dataset. 
For \modelnamesdf{}, we train a separate model on each subset $D_i \in D$ and evaluate its performance on the corresponding test subset.

\subsection{Experimental Setup}

We train \modelname{} in three different sizes: base (250M parameters), large (800M parameters), and extra-large (3B parameters). 
The models are fine-tuned using the pre-trained Flan-T5 checkpoint, with the number of parameters determined by the model size. 
We experiment with both LoRA and full fine-tuning (FT) to evaluate the model's performance under different training settings. 
In the case of LoRA, we use adapter layers for all fully connected layers available in the corresponding models.
In all cases, we used the AdamW optimizer \cite{loshchilov2018decoupled} and an effective batch size of 16.

When performing full fine-tuning, we use a maximum learning rate of $5 \times 10^{-5}$. 
For LoRA adaptation, we use a higher learning rate of $1 \times 10^{-4}$ \cite{hu2021lora}.
In both cases, we use a linear learning rate scheduler with a warm-up phase. 
Specifically, the learning rate is linearly increased from zero to the maximum value over the first 10\% of the total training steps (warm-up phase). After the warm-up phase, the learning rate is linearly decreased back to zero for the remainder of the training. 
For \modelnamecdf{}, we set the maximum number of training epochs to 2, given the higher amount of training data and computational resources required for training on the combined dataset.
For \modelnamesdf{}, we train each model for up to 10 epochs, following established best practices.
In all cases, the best model is selected based on the validation WER and used for evaluation on the test subsets of the HyPoradise dataset.
All experiments are conducted on 2 NVIDIA A100 GPUs with 80GB of memory each. 
We use the Hugging Face Transformers library \cite{wolf-etal-2020-transformers} for model training and evaluation.

\noindent
\textbf{Evaluation Metrics:} We evaluate the models using the Word Error Rate (WER) metric, which measures the percentage of words in the model's output that differ from the reference transcription. 
We follow the implementation provided in the original HyPoradise repository\footnote{\url{https://github.com/Hypotheses-Paradise/Hypo2Trans}} to have consistent results.

\begin{table*}[]
\centering

\resizebox{\textwidth}{!}{%
\begin{tabular}{@{}c|c|c|c|cccccc|cccccc@{}}
\toprule
\multirow{3}{*}{Dataset} & \multirow{3}{*}{Baseline} & \multirow{3}{*}{\begin{tabular}[c]{@{}c@{}}FlanT5\\ ZS\end{tabular}} & \multirow{3}{*}{\begin{tabular}[c]{@{}c@{}}ChatGPT-3.5\\ ICL\end{tabular}} & \multicolumn{6}{c|}{\colorbox{lightblue}{\modelnamesdf{}}}     & \multicolumn{6}{c}{\colorbox{lightred}{\modelnamecdf{}}}    \\ \cmidrule(l){5-16} 
                         &                           &                                                                      &                                                                            & \multicolumn{3}{c}{LoRA} & \multicolumn{3}{c|}{Finetuning}     & \multicolumn{3}{c}{LoRA}   & \multicolumn{3}{c}{Finetuning} \\
                         &                           &                                                                      &                                                                            & 250M    & 800M   & 3B    & 250M & 800M         & 3B            & 250M & 800M & 3B           & 250M  & 800M  & 3B             \\ \midrule
WSJ                      & 4.5                       & 6.8                                                                  & 4.9                                                                        & 2.8     & 2.8    & 2.6   & 2.7  & 2.6          & 2.8           & 2.7  & 2.6  & 2.4          & 2.6   & 2.5   & \textbf{2.3}   \\
ATIS                     & 8.3                       & 18.6                                                                 & 3.7                                                                        & 1.3     & 3.2    & 1.4   & 1.7  & 1.5          & \textbf{1.1}  & 2.7  & 3.0  & 2.1          & 2.3   & 1.3   & \textbf{1.1}   \\
CHiME-4                  & 11.1                      & 11.8                                                                 & 10.7                                                                       & 6.7     & 6.7    & 5.3   & 6.9  & 5.9          & 5.5           & 5.5  & 4.8  & 4.9          & 5.0   & 4.5   & \textbf{3.9}   \\
Tedlium-3                & 8.5                       & 9.9                                                                  & 5.4                                                                        & 4.6     & 4.7    & 4.7   & 4.7  & 4.6          & 4.8           & 4.5  & 4.4  & \textbf{4.0} & 4.3   & 4.3   & 4.2            \\
CV-accent                & 14.8                      & 18.4                                                                 & 13.6                                                                       & 13.4    & 12.7   & 11.6  & 13.3 & 11.7         & \textbf{10.8} & 13.6 & 12.4 & 11.7         & 13.7  & 12.2  & 11.1           \\
SwitchBoard              & 15.7                      & 33.4                                                                 & 16.4                                                                       & 15.0    & 15.0   & 14.6  & 17.2 & 15.2         & 16.6          & 17.1 & 17.0 & 15.2         & 16.6  & 14.7  & \textbf{14.1}  \\
LRS2                     & 10.1                      & 42.3                                                                 & 11.0                                                                       & 9.1     & 9.0    & 8.6   & 9.3  & \textbf{8.4} & \textbf{8.4}  & 10.0 & 9.3  & 9.5          & 10.6  & 9.6   & 8.7            \\
CORAAL                   & \textbf{21.4}             & 25.7                                                                 & 24.8                                                                       & 31.2    & 23.5   & 23.5  & 26.6 & 22.2         & 23.5          & 23.2 & 22.5 & 22.9         & 22.9  & 22.4  & 22.2           \\ \midrule
Average                  & 11.8                      & 20.9                                                                 & 11.3                                                                       & 10.5    & 9.7    & 9.0   & 10.3 & 9.0          & 9.2           & 9.9  & 9.5  & 9.1          & 9.8   & 8.9   & \textbf{8.5}   \\ \bottomrule
\end{tabular}%
}

\caption{Comparison of WER results for \colorbox{lightblue}{\modelnamesdf{}} (Single Dataset, SD) and \colorbox{lightred}{\modelnamecdf{}} (Cumulative Dataset, CD). In the SD setting, each model is trained on a single subset of the HyPoradise dataset. In the CD setting, one model is trained on the combined subsets of the HyPoradise dataset. The models were adapted using both LoRA and full fine-tuning (FT) methods at different parameter scales. The best overall results are highlighted in \textbf{bold}. 
The last row reports the average across all datasets.
}
\label{tab:overall_results}
\end{table*}

\section{Results and Discussion}
\label{sec:results}

In this section, we present and analyze the performance of \modelname{} on the HyPoradise dataset, focusing on post-ASR error correction across different ASR domains. 
The results are summarized in Table \ref{tab:overall_results}, where we compare the WER scores of \modelname{} under different training settings, model sizes, and fine-tuning strategies. 
For an overall comparison, we have included the average WER across all datasets in the last row of Table~\ref{tab:overall_results}.
For a comprehensive evaluation, we also incorporate baseline results from the original HyPoradise \cite{chen2024hyporadise} experiments, which are obtained using ASR without error correction. 
In addition, we evaluate two LLM-based approaches without fine-tuning: (1) Flan-T5 in a zero-shot (ZS) setting, using the 3B parameter model provided with the target sample and its 5-best hypotheses, and (2) in-context learning (ICL) using ChatGPT (\texttt{gpt-3.5-turbo-0125}), where the model also receives 5 training examples, each containing n-best lists and transcriptions. 
These training examples are selected based on the similarity of their first hypotheses to the target sample.
We did not evaluate Flan-T5 in the ICL setting because prior research \cite{t5_cannot_ICL} has shown that encoder-decoder models struggle with ICL and even exhibit decreased performance when provided with in-context examples.
These general-purpose LLM-based approaches, without task-specific adaptation, provide a broader context for evaluating post-ASR correction performance.

\vspace{1mm}
\noindent
\textbf{Single Dataset Models (SD):}
\modelnamesdf{} results indicate that scaling up the model size generally leads to improvements in WER. For instance, the 3B parameter model achieves the best performance on multiple datasets such as ATIS, CHiME-4, and CV-accent, with significant WER reductions compared to smaller models. 
The base model (250M parameters) trained with LoRA achieves better WER scores than the FT counterpart on the majority of subsets. This may be due to the base model's limited capacity, which can lead to overfitting when fine-tuning all model parameters using only a specific subset of data. 
For example, on SwitchBoard, the LoRA-adapted base model achieves a WER of 15.0\%, compared to 17.2\% for the fully fine-tuned counterpart.

\vspace{1mm}
\noindent
\textbf{Cumulative Dataset Models (CD):}
The 3B \modelnamecdf{} model with full fine-tuning achieves the best overall performance in 4 out of 8 datasets, with significant WER reductions in domains such as WSJ, CHiME-4, and SwitchBoard. 
This indicates that certain domains may benefit from training on a cumulative dataset, leveraging the diverse linguistic patterns and contexts provided by multiple ASR domains.
Consistent with SD results, full fine-tuning (FT) outperforms LoRA across all datasets. 
This is likely due to the scale of training data, which helps mitigate overfitting and allows the model to fully exploit its parameter capacity.
While larger models generally deliver better performance, the 800M model often provides the best tradeoff between performance and computational complexity. It achieves substantial improvements over the base model while maintaining a lower computational overhead compared to the 3B model.

\vspace{1mm}
\noindent
\textbf{LoRA vs. Full Fine-tuning:}
Comparing LoRA and full fine-tuning (FT) strategies, FT consistently shows better performance across most datasets. 
The only exception occurs at smaller model sizes in the SD setting, where LoRA outperforms FT on the majority of the subsets, possibly because the limited model capacity leads to overfitting when all parameters are fine-tuned. 
LoRA mitigates this by adapting only a subset of parameters, reducing the risk of overfitting and task forgetting, especially in low-resource scenarios. 
However, most of the considered subsets contain around 4,000 samples or more, except for CORAAL, which has approximately 2,000 training examples.
The results highlight that while LoRA is computationally efficient, GenSEC tasks significantly benefit from full model fine-tuning. 
In the CD settings, the 3B \modelname{} FT model achieves an average improvement of $0.64\pm0.46$ absolute WER points compared to its LoRA counterpart. 
This highlights the advantages of full fine-tuning, especially as the scale of training datasets increases. 

\vspace{1mm}
\noindent
\textbf{Model Size and Training Strategy:}
Our results clearly show that scaling up the model size and training on a combination of all training subsets with full fine-tuning can enhance the model's accuracy in post-ASR error correction. 
The 3B \modelnamecdf{} model stands out as the best-performing model across several subsets.
The diverse nature of the HyPoradise dataset allows us to evaluate the generalization capabilities of \modelname{} across various ASR domains. 
The results indicate that \modelnamecdf{}  demonstrates robust generalization across different types of ASR data, from noisy environments (CHiME-4) to conversational speech (SwitchBoard).

However, consistent with previous findings \cite{chen2024hyporadise}, the model struggles with the CORAAL subset, with no model achieving a WER below the baseline.
This aligns with existing research \cite{koenecke2020racial, wassink2022uneven, robertson2024quantifying}, which identifies acoustic modeling as the primary challenge for African-American language in ASR systems. 
As shown in Table \ref{tab:stats_tokens}, this dataset has the highest percentage of tokens in the reference transcriptions that are absent in any of the $n$-best hypotheses, with more than 90\% of references lacking an exact match with any of the $n$-best hypotheses, likely contributing to its difficulty. 
Although larger models seem to mitigate the issue to some extent, the results suggest that further research is needed to address the challenges posed by this specific subset.
Those findings emphasize the effectiveness of larger models and cumulative training datasets in enhancing post-ASR error correction, while also highlighting areas for improvement in handling highly challenging subsets like CORAAL.

\vspace{1mm}
\noindent
\textbf{Comparison with Additional Baselines:}
When evaluating LLM-based approaches without fine-tuning, we observe distinct patterns. 
First, Flan-T5 in the zero-shot (ZS) setting does not improve upon the baseline results. 
Although Flan-T5 is instruction-tuned, it struggles to generalize to the post-ASR correction task without any task-specific tuning, highlighting the limitations of using a model in a zero-shot configuration for this task.
On the other hand, in-context learning (ICL) with ChatGPT performs better than the baseline in 4 out of 8 cases, demonstrating the generalization capabilities of LLMs like ChatGPT, even when only provided with similar training examples rather than explicit fine-tuning.
However, it is important to note that neither Flan-T5 (ZS) nor ChatGPT with ICL are specifically trained for this task; they are simply used as general-purpose LLMs applied to post-ASR correction.
In contrast, \modelname{} models, which are explicitly fine-tuned for the task, consistently outperform both Flan-T5 (ZS) and ChatGPT (ICL), demonstrating the need of task-specific adaptation to achieve optimal performance in this context.

\section{Conclusions}
\label{sec:conclusions}

We introduced \modelname{}, an encoder-decoder model based on Flan-T5, for post-ASR error correction. 
Extensive experiments on the HyPoradise dataset demonstrate that \modelnamecdf{}, trained on a combined dataset, generally outperforms single dataset models, especially when fully fine-tuned. 
This suggests the potential for creating a general-purpose model capable of effective error correction across various ASR domains. 
Larger models showed significant performance gains, highlighting the importance of model scaling and diverse training data. 
While LoRA offers efficiency, full fine-tuning provides, on average, better results. 
Despite improvements, challenges with datasets like CORAAL remain, indicating areas for further research. 
We plan to expand our analysis to larger models (e.g., 11B parameters) to fully understand the impact of scaling and to further develop a versatile post-ASR correction model.

\bibliographystyle{IEEEbib}
\bibliography{strings}

\end{document}